\documentclass[conference]{IEEEtran}
\IEEEoverridecommandlockouts
\usepackage{cite}
\usepackage{amsmath,amssymb,amsfonts}
\usepackage{algorithmic}
\usepackage{graphicx}
\usepackage{textcomp}
\usepackage{xcolor}
\usepackage{subcaption}
\usepackage{float}

\usepackage{bm}
\usepackage{tikz}
\usepackage{amsmath}
\usepackage{color}
\usepackage{algorithm}
\usepackage{algorithmic}
\usetikzlibrary{intersections,chains,scopes,fit,trees,quotes,calc,fadings,positioning,calc,patterns}

\tikzset{
every path/.style = {
->,
> = stealth,
very thick,
rounded corners},
state/.style = {
rectangle,
rounded corners,
draw=black,
fill=gray!10,
thick,
minimum height=1em,
inner sep=5pt,
text centered
},
note/.style = {
fill = yellow,
text width = 3.2cm,
text centered,
},
node distance=2cm and 1cm,
}

\usepackage[utf8]{inputenc}

\DeclareFixedFont{\ttb}{T1}{txtt}{bx}{n}{12} 
\DeclareFixedFont{\ttm}{T1}{txtt}{m}{n}{12}  

\usepackage{color}
\definecolor{deepblue}{rgb}{0,0,0.5}
\definecolor{deepred}{rgb}{0.6,0,0}
\definecolor{deepgreen}{rgb}{0,0.5,0}

\usepackage{listings}

\newcommand\pythonstyle{\lstset{
		language=Python,
		basicstyle=\ttfamily,
		otherkeywords={self},             
		keywordstyle=\ttb\color{deepblue},
		emph={MyClass,__init__},          
		emphstyle=\ttb\color{deepred},    
		stringstyle=\color{deepgreen},
		frame=tb,                         
		showstringspaces=false            %
}}

\lstnewenvironment{python}[1][]
{
	\pythonstyle
	\lstset{#1}
}
{}

\newcommand\pythoninline[1]{{\pythonstyle\lstinline!#1!}}

\usepackage{amsmath}

\newcommand{\eqn}[2]
{
	\begin{equation}#1
	#2
	\end{equation}
}

\newcommand{\eqna}[1]
{
	\begin{eqnarray}
	#1
	\end{eqnarray}
}

\newcommand{\fr}
{ \FOR}

\newcommand{\efr}
{ \ENDFOR}

\newcommand{\st}
{\STATE}

\newcommand{\rt}
{\RETURN}

\newcommand{\im}[1]
{
	\begin{itemize}
		#1
	\end{itemize}
}

\def\BibTeX{{\rm B\kern-.05em{\sc i\kern-.025em b}\kern-.08em
    T\kern-.1667em\lower.7ex\hbox{E}\kern-.125emX}}
\begin{document}



\title{
Context Does Matter: End-to-end Panoptic Narrative Grounding with Deformable Attention Refined Matching Network


{\footnotesize }
\thanks{This work was partially supported by “Qing Lan Project” in Jiangsu universities, NSFC under No. 62106081, 62376113, 92370119, RDF with No. RDF-22-01-020, Jiangsu Science and Technology Programme under No. BE2020006-4 and Natural Science Foundation of the Jiangsu Higher Education Institutions of
China under No. 22KJB520039}
}

\author{
\IEEEauthorblockN{Yiming Lin, Xiao-Bo Jin$^\dagger$, Qiufeng Wang}
\IEEEauthorblockA{
\textit{School of Advanced Technology} \\
\textit{Xi'an Jiaotong-Liverpool University, Suzhou, China}\\
\{yiming.lin21@student, xiaobo.jin@, qiufeng.wang@\}xjtlu.edu.cn}
\and
\IEEEauthorblockN{Kaizhu Huang}
\IEEEauthorblockA{
\textit{Data Science Research Center} \\
\textit{Duke Kunshan University, Suzhou, China}\\
kaizhu.huang@dukekunshan.edu.cn}
}

\maketitle

\begin{abstract}
Panoramic Narrative Grounding (PNG) is an emerging visual grounding task that aims to segment visual objects in images based on dense narrative captions. The current state-of-the-art methods first refine the representation of phrase by aggregating the most similar $k$ image pixels, and then match the refined  text representations with the pixels of the image feature map to generate segmentation results. However, simply aggregating sampled image features ignores the contextual information, which can lead to phrase-to-pixel mis-match. In this paper, we propose a novel learning framework called Deformable Attention Refined Matching Network (DRMN), whose main idea is to bring deformable attention in the iterative process of feature learning to incorporate essential context information of different scales of pixels. DRMN iteratively re-encodes pixels with the deformable attention network after updating the feature representation of the top-$k$ most similar pixels. As such,  DRMN can lead to  accurate yet discriminative pixel representations, purify the top-$k$ most similar pixels, and consequently alleviate the  phrase-to-pixel mis-match substantially.
Experimental results show that our novel design significantly improves the matching results between text phrases and image pixels. Concretely, DRMN achieves new state-of-the-art performance on the PNG benchmark with an average recall improvement 3.5\%. The codes are available in: https://github.com/JaMesLiMers/DRMN.

\end{abstract}


\section{Introduction}

\begin{figure}[htbp]
\centerline{\includegraphics[width=0.95\linewidth]{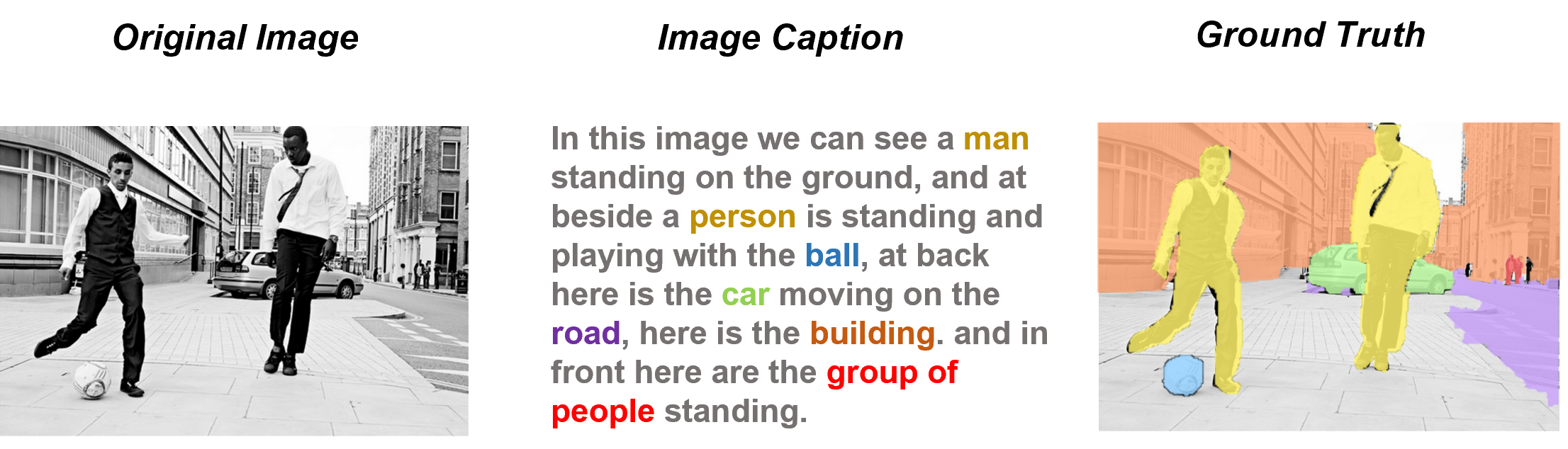}}
\caption{Illustration of the PNG problem: Given an image (left) and corresponding caption (middle), the goal is to generate a panoptic segmentation (right) based on all visual objects contained in the caption (i.e., labeling each object and its associated segmented region  with the same color).}
\label{png_problem_def}
\end{figure}

Panoptic Narrative Grounding (PNG) \cite{gonzalez2021panoptic}, one emerging visual grounding task, has recently drawn great attention in data mining and computer vision including grounded context recognition \cite{pratt2020grounded}, visual question answering \cite{hudson2019gqa}, and visual-language model pre-training \cite{li2022grounded}. Given an image and its associated dense narrative caption, the goal of PNG  is to segment the visuals of things and stuff based on the visuals mentioned in the caption (see illustration in  Fig.~\ref{png_problem_def}). In contrast to other related tasks,  PNG extends the grounding range from the bounding box of the foreground class (called ``object") to a segmentation mask containing both foreground and background classes (named ``object" and ``Stuff"), thus defining the finest-grained alignment between multiple noun phrases and segments. 
 A detailed comparison between  PNG and other related vision-based tasks can be seen in Sect.~\ref{Sec:Related}. 

\begin{figure*}[htbp]
\centerline{\includegraphics[width=0.95\linewidth]{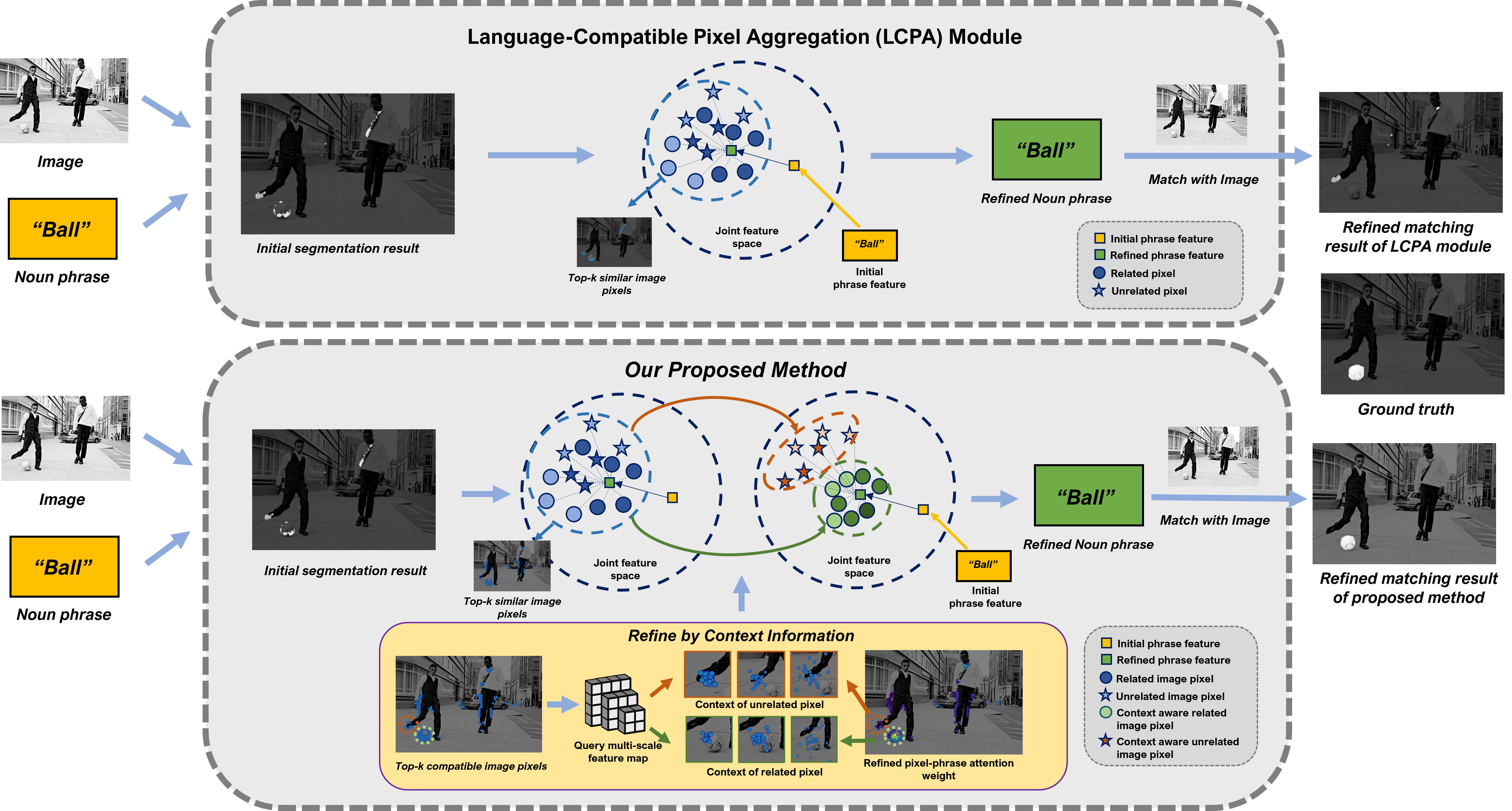}}
 \caption{Insight of our proposed method. The upper part addresses the limitation of  LCPA with a hard example. We introduce the essential context information in the multi-scale feature map as a cue to refine the sampled top-$k$ image feature. By sharpening the representation of different visual objects points, our method  can filter out the sampled points by purifying unrelevant visual objects, which further enhances the final segmentation result.}
\label{problem_example}
\end{figure*}


In general, there are two families of methods for PNG. The first type of methods typically exploits a two-stage pipeline~\cite{gonzalez2021panoptic}, which matches by computing the affinity matrix between object proposals (extracted by off-the-shelf models) and noun phrases. As such, the object proposal model, i.e., the off-the-shell model will limit the performance ceiling. On the other hand, the one-stage or end-to-end  methods~\cite{ding2022ppmn, wang2023towards} alleviate this problem by directly generating a response map between all noun phrases and  image pixels. To better fuse information from different modalities, Ding et al. \cite{ding2022ppmn} propose a language-compatible pixel aggregation (LCPA) module to aggregate the most compatible features from images to noun phrases. Namely, taking each noun phrase as a query feature,  LCPA samples top-$k$ image-compatible features, which are then used as key and value features. Finally the multi-head cross-modal attention is adopted to aggregate visual features.




Albeit its promising performance, LCPA simply  aggregates sampled image features without taking into account the contextual information, which could lead to serious phrase-to-pixel mis-match. Concretely, LCPA tends to push the phrase feature towards the center of top-$k$ sampled image features. This algorithm works well when the top-$k$ sampled features are related to the target visual object with high similarity. However, such strategy may also inevitably introduce unrelated image pixels. As illustrated in a hard example of Fig.~\ref{problem_example}, the unrelated object's pixels (for ``feet") fused with the related object's pixels ( for ``ball") dominate the top-$k$ sampled features with high similarity, thus inducing a serious mis-match. To alleviate this problem, we argue that  relevant context information is crucial for differentiating and purifying the top-$k$ sampled points.  Essentially, integration of context information would enable  more accurate and discriminate pixel representation, since related pixels enjoy more similar context whilst unrelated ones may share distinctive context. In other words, integration of context information could sharpen the representation of different visual pixels, thereby providing potentials to improve the segmentation performance.

Motivated from the above observations and inspired by the current object detection method \cite{zhu2020deformable},  we design a novel deformable attention mechanism  to extract essential pixel contextual information in multi-scale feature maps in an iterative way, resulting in a simple yet effective end-to-end model, called Deformable Attention Fine Matching Network (DRMN). An overview of its framework is shown in Fig. \ref{model_overview}. Similar to Transformer, DRMN is a multi-scale encoding-decoding method that offers a full range of context information at different scales, including global information and local information. For feature extraction, we engage the deformable attention to encode image features at different levels, which will generate a cross-fused multi-scale word-pixel matching matrix to obtain initial word-pixel matching results. In addition, in the feature aggregation stage, we further incorporate word embedding representations into pixel encoding representations. Specifically, we follow insights from the DETR decoder~\cite{kamath2021mdetr}  to refine the sampled features: for each word vector, our model queries its nearest $k$ pixels and applies an attention mechanism to encode them after updating the vector representations of these pixels with the word vector.

Our contributions are four-fold:

\im{

\item From the perspective of multimodal information fusion, we design a deformable attention model with multi-scale encoding and decoding functions in the aggregation process of pixel features to better encode the context information around the pixel, effectively alleviating the phrase-to-pixel mis-match problem.

\item {From the perspective of model structure, unlike the existing DETR-based models \cite{kamath2021mdetr,shi2022dynamic}, our novel design leverages directly  multimodal transformers for highly intertwined feature aggregation. Inheriting insights from DETR feature aggregation, our new approach offers a more sparse and interpretable way of exploiting DETR for vision-based tasks.}

\item From an algorithmic perspective, we simplify the multi-round pixel feature refinement process into an iterative process of two subproblems: the fuzzy K-means clustering subproblem and the multi-objective assignment subproblem, the latter of which can be efficiently solved by online gradient descent.

\item From an experimental point of view, the results of multiple categories and the overall results on the public PNG benchmark show the superiority of our method compared to previous methods, where the average recall rate  is 3.5\% higher than the second-ranked method.

}

\section{Related Work}\label{Sec:Related}

In this section, we overview  PNG in contrast to different related vision-based tasks. Overall, Table \ref{tab:related-tasks} shows the comparison granularities of related
vision-based tasks, among which the PNG task provides the most fine-grained alignment between different types of nouns and segmentation. 
\begin{table}[htbp]
\caption{Comparison of different granularities of related vision-based tasks. Considering the typical segmentation categories in computer vision tasks between things (countable objects) and stuff (amorphous regions of similar texture), the datasets of the other tasks mainly focus on things categories. }
\begin{center}
\begin{tabular}{|l|c|c|c|}
\hline
\textbf{Grounding}&{\textbf{Language}}&{\textbf{Visual}}&{\textbf{Semantic}} \\
\textbf{Task} & \textbf{\textit{Granularity}}& \textbf{\textit{Granularity}}& \textbf{\textit{Generality}}\\
\hline
REC \cite{de2017guesswhat} & Short phrase & Bounding box & Things  \\
\hline
PG \cite{plummer2015flickr30k} & Noun phrase & Bounding box & Mainly Things  \\
\hline
RES \cite{hu2016segmentation} & Short phrase & Segmentation & Things  \\
\hline
PNG \cite{gonzalez2021panoptic} & Noun phrase & Segmentation & Things + Stuff  \\
\hline
\end{tabular}
\label{tab:related-tasks}
\end{center}
\end{table}
\subsection{Visual Grounding with Bounding Box Regression}
The goal of Referent Expression Comprehension (REC) task is to predict the corresponding bounding box in an image for a given referring expression. Current methods can be categorized into two-stage and one-stage approaches. Two-stage methods \cite{zhang2018grounding} first propose bounding box proposals in the image,  then match the proposal-referring expression pairs. Inspired by object detection techniques \cite{girshick2015fast}, the one-stage methods~\cite{yang2019fast} directly generate results based on the input textual information without explicit matching. Recently, some methods have explored multi-modal pre-training models \cite{chen2020uniter} in REC~\cite{kazemzadeh2014referitgame}, taking architectures similar to BERT, to obtain joint representations of images and texts.

The phrase grounding task aims to find the corresponding bounding box in an image for multiple noun phrases mentioned in an input caption. Early methods \cite{plummer2015flickr30k,plummer2017phrase} adopted representation learning, which first project region proposal and phrase embeddings onto a same subspace, then learn semantic similarity between them. In recent years, researchers have explored various methods~\cite{dogan2019neural, mu2021disentangled,yu2020cross} for fusing and learning multi-modal features. It is worth mentioning that recent large-scale visual-language pre-training models \cite{liu2023grounding, zhang2022glipv2} have adopted weakly supervised phrase grounding \cite{wang2020maf} loss to align image-noun phrase pairs.


Recently, some methods \cite{deng2021transvg, kamath2021mdetr, shi2022dynamic} have modified the transformer-based object detection framework to address the aforementioned bounding box regression problems. TransVG~\cite{deng2021transvg} first proposed a pure transformer framework for visual grounding tasks. Furthermore, some methods \cite{kamath2021mdetr,shi2022dynamic} drew inspirations from the DETR object detection framework. MDETR \cite{kamath2021mdetr} employed a transformer encoder-decoder structure, where the transformer simultaneously extracted features from both the image and text in the encoder, and introduced QA-specific queries in the decoder for visual grounding-related task decoding. Dynamic MDETR \cite{shi2022dynamic} engaged the idea of deformable attention in the decoder to reduce computation. It is worth noting that our approach differs from the aforementioned methods. Instead of using a transformer to simultaneously encode image and text information, we handle the multi-modal feature interaction in the decoder through top-$k$ sampling. In particular, in the decoder, we consider the features of the top-$k$ image positions as object queries, and design the deformable attention mechanism to extract object-relevant features and then aggregate the extracted object query features into the textual features.


\subsection{Visual Grounding with Segmentation}


The task of Referent Expression Segmentation (RES) is to generate a segmentation map of the referred object according to the input referring expression. The first proposed method \cite{hu2016segmentation} on this task is a one-stage model that first concatenated textual features and global image feature, then decoded the segmentation mask through deconvolution layers. Recently, inspired by multi-modal transformers, various fine-grained modeling approaches \cite{ding2021vision,yang2022lavt} have been proposed to facilitate interactions between different modalities. For example, SHNET \cite{jain2021comprehensive} concatenated textual features with different levels of image features as joint input of the transformer, then adopted language features to guide the information exchange between different levels of image features. The LAVT model \cite{yang2022lavt} developed the PWAM module, which directly used attention to expand textual information to the size of the image feature map for pixel-word feature fusion.


PNG aims to segment corresponding things or stuff in an image based on the multiple noun phrases mentioned in the image caption. This task was initially proposed with a two-stage method by González et al.~\cite{gonzalez2021panoptic}, along with a dataset. They extracted segmentation proposals from off-the-shelf models which were matched with the extracted noun phrase features. Later, some work explored the one-stage  paradigm. For example,  PPMN~\cite{ding2022ppmn} achieved feature fusion between different modalities through a sampling strategy end-to-end. EPNG \cite{wang2023towards} further optimized the inference speed and achieved real-time segmentation effects while sacrificing little accuracy. 

\begin{figure*}[htbp]
\centerline{\includegraphics[width=.95\linewidth]{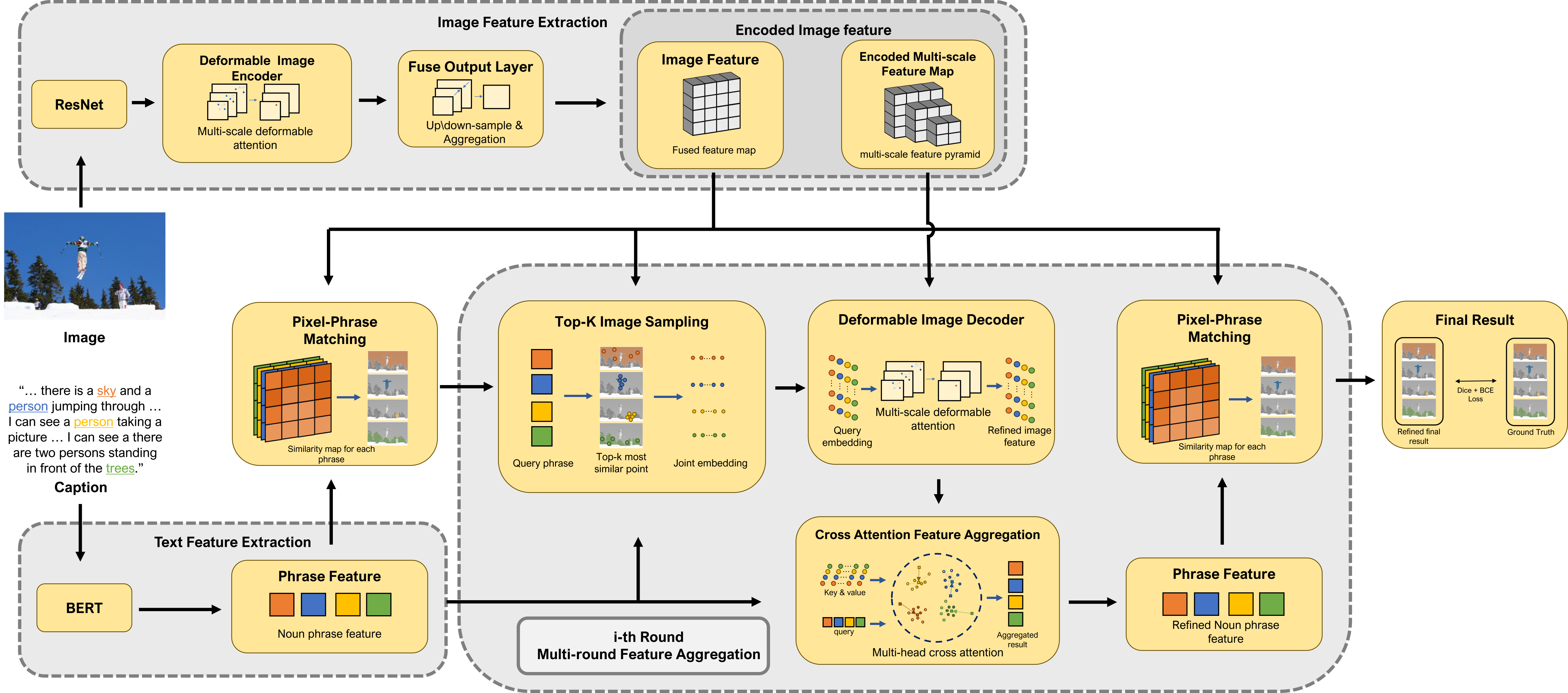}}
\caption{Overview of our model. {We integrate essential image context information in feature extraction and multi-round feature aggregation phases with deformable attention. First, we utilize BERT to encode textual features and employ deformable attention to encode multi-scale image feature maps. Furthermore, we generate initial image-text matching results based on textual and image features. Finally, in the multi-round feature aggregation, we aggregate the top-$k$ image features into text feature based on the matching results. The model utilizes deformable attention to refine sampled image features further, then aggregates the refined features into textual features through a cross-attention mechanism to generate improved matching results.}}
\label{model_overview}
\end{figure*}


\section{Main Method}
In this section, we first introduce the process of feature extraction for image text (\ref{feature extraction}). Then, we describe how initial segmentation results are generated (\ref{model matching}). Subsequently, we present our proposed Multi-round Visual-Language Aggregation Module, which selectively aggregates image features into textual features to enhance the model's performance (\ref{fuse model}). Finally, we detail the loss function and introduce the training process of the model (\ref{model loss}). The overall workflow of our model is shown in Fig.~\ref{model_overview} and the pseudo code of the algorithm is shown in Alg. 1.  

\subsection{Feature Extraction}
\label{feature extraction}


In the feature extraction stage, we can employ off-the-shelf methods to extract features from visual and linguistic modalities.


For the text modality, we leverage BERT to extract features for each word in the image caption. Specifically, we focus on extracting features from the noun phrase part $T$ of the title
$\mathcal{G} = \textrm{BERT}(T) \in R^{n \times d}$, 
where $n$ represents the maximum number of words in all input noun phrases and $d$ denotes the dimensionality of the textual feature embedding representation.


As for image modality, given any image $I$, we use ResNet as the backbone to extract multi-scale feature pyramid $S = \{F_2,F_3,F_4,F_5\}$ such as 
\begin{equation}
 F_{l} = \textrm{flatten}(\textrm{ResNet}(I,l)),\quad l = 2,3,4,5.
\end{equation}
Here $I$ is an RGB image with height $h$ and width $w$ and the $l$-th scale output will be a matrix of size $\frac{hw}{4^l} \times c$, where $c$ is the number of channels in the output map.

In the subsequent feature extraction stage, first we normalize the coordinates $x$'s and $y$'s of all points in the feature map of each scale to the range $[0,1]$. We then grid them to obtain the reference point matrix with the same size
$p_l = \textrm{flatten}(\textrm{grid}(M_l)) \in R^{\frac{hw}{4^l} \times 2}$, 
where the size of the tensor $M$ is $\frac{h}{2^l} \times \frac{w}{2^l} \times 2$ and its elements $M_{l}(i,j) = (i,j)$. 


Meanwhile, we add the feature maps of multiple scales obtained by the feature pyramid and their position codes and straighten them to get the matrix of $\frac{hw}{4^{l}} \times c$ 
\begin{equation}
\hat{F}_l = F_l + \textrm{pos}(F_l), \quad l = 2,3,4,5,
\end{equation}
where $\textrm{pos}(\cdot)$ represents the positional encoding function. We take these feature representations and concatenate them row by row into a matrix as the input of the DeformLayer function in the initial stage
\eqn{}{
\hat{F} = \textrm{catrow}(\{\hat{F}_l\}).
}

\subsection{Deformable Layer}

In order to better integrate feature information at different scales, we use multiple deformable attention layers to aggregate information from various levels of the feature pyramid according to the position and characteristics of the input feature points, where the input of deformable attention is the query features $Q = \hat{F}$,  the reference positions $P = \{p_2,p_3,p_4,p_5 \}$ corresponding to the feature point in the image and multi-scale feature map $S = \{F_2,F_3,F_4,F_5\}$.


Similar to Transformer, on multiple feature maps of the pyramid, each feature point will be re-expressed as a linear combination of other feature points. The difference is that its value $V$ is based on the feature map with a bilinear sampling operation instead of itself. Moreover,  both the sampling offset and the self-attention correlation coefficient depend on the query $Q$, specifically described as follows
\begin{eqnarray}
    \Delta p_l & = & \hat{F} W_l^{p} \\
    V_l & = & \phi_{\textrm{bilin}}(F_l,p_l + \Delta p_l) \\
    V & = & \textrm{catrow}(\{V_l\})  \\
    \hat{V} & = & \textrm{softmax}(\hat{F} W^a) V W^{v},
\end{eqnarray}
where $l = 2,3,4,5$ means transformation on multiple scales, $W_l^p$, $W^v$ and $W^a$ represent the linear mapping to be learned, and $\phi_{\textrm{bilin}}(F_l,p_l + \Delta p_l)$ indicates that pixels are sampled by bilinear interpolation on the position $p_l + \Delta p_l$ of the feature map $F_l$.  Similarly, multiple heads are introduced to obtain image feature representations with multiple attentions, and these representations are concatenated and linearly mapped to obtain a multi-scale refined representation $\mathcal{V}$.

Subsequently, we can construct a deformable coding layer with deformable attention representation
\begin{equation}
    \mathcal{F} = \textrm{FFN}(\textrm{norm}(\mathcal{V} + \textrm{dropout}(\mathcal{V}))),
\end{equation}
where FFN, norm and dropout indicate feedforward network layer,  normalization layer, and dropout layer respectively.

For convenience, we represent the above whole process as the following function
\begin{equation}
    \mathcal{F} = \textrm{DeformLayer}(\hat{F},\{F_l\},\{p_l\}).
\end{equation}

\subsection{Image and Text Matching}
\label{model matching}

Below we describe how to match text embedding $\mathcal{G}$ with multiple feature maps $\mathcal{F}_l$ of different scales to obtain multiple similarity matrices. 

In the initial stage, we concatenate multi-scale feature maps by row as DeformLayer function to get their attention representation, and then we restore them into multiple feature maps $\mathcal{F}_l$ (line 8-12 in Alg. 1). Then, using the third layer as a benchmark, all the other layers are down-sampled or up-sampled to the same size as the feature map of the third layer
\begin{equation}
    \bar{F}_l  = 
        \phi_{\textrm{sampling}}(\textrm{reshape}(\mathcal{F}_l),2^{l - 3}),\quad l = 2,3,4,5.
\end{equation}
In this way, we can fuse the feature map output with the same scale, and convert the fused image output into  vectors to facilitate matching with the text vector (line 13 in Alg. 1)
\begin{equation}
    \mathbb{F} = \textrm{vect}\left(\frac{1}{4} \sum_{l = 2}^5 \bar{F}_l  \right),
\end{equation}
where $\textrm{vect}(\cdot)$ means pulling a three-dimensional tensor $\mathbb{F}\in R^{(h/8) \times (w/8) \times c}$ into a two-dimensional vector $\mathbb{F} \in R^{(hw/8^2) \times c}$. 

Now, we map the representation $G$ of phrases to the feature space of pixels to compute their similarity matrix (line 15 in Alg. 1)
\begin{eqnarray}
    \mathcal{\hat{G}} = \mathcal{G} V^g, \quad
    H = \textrm{sigmoid}\left(\hat{\mathcal{G}} \mathbb{F}^T \right),
\end{eqnarray}
where $V^g$ is a projection matrix and $\textrm{sigmoid}$ is the sigmoid function. 

It is worth noting that during the iterative process, we directly use the latest representation $\hat{F}$ of the pixel to calculate the similarity matrix (line 25 in Alg. 1)
\eqn{}{
 H = \textrm{sigmoid}\left(\hat{\mathcal{G}} \hat{F}^T \right).
}

\subsection{Multi-round Feature Aggregation Module}
\label{fuse model}










\begin{algorithm}
\caption{Multi-round Feature Aggregation}
\label{alg:multi-round-aggregation}
\begin{algorithmic}[1]
\st Input: Image $I$ and caption $T$

\st $\mathcal{G} = \textrm{BERT}(T)$
\fr {$l = 2,3,4,5$}
\st $p_l = \textrm{flatten}(\textrm{grid}(M_l))$
\st $F_l = \textrm{flatten}(\textrm{ResNet}(I,l))$
\st $\hat{F} = F_l + \textrm{pos}(F_l)$
\efr 

\st $\hat{F} = \textrm{catrow}(\{\hat{F}_l\})$

\fr {$t = 1,2,\cdots,T$}
\st $\hat{F} = \textrm{DeformLayer}(\hat{F},\{F_l\},\{p_l\})$
\efr

\st $\{\mathcal{F}_l\} = \textrm{splitrow}(\hat{F})$

\st $\hat{F} = \textrm{avg}(\{\mathcal{F}_l\})$

\st $\mathcal{\hat{G}} = \mathcal{G} V^{g}$

\st $H = \textrm{sigmoid}(\mathcal{\hat{G}}\hat{F}^T)$

\st $\mathcal{H} = []$
\fr {$i = 1,2,\cdots,I$}

\st $S = \textrm{topk}(H,k)$

\fr {$j = 1,2,\cdots,n$}
\st $s = S[j,:]$

\st $\hat{F}[s,:] = \hat{F}[s,:] +  \textrm{pos}(\hat{F}[s,:]) + \hat{\mathcal{G}}[j,:]$
\st $\hat{F}[s,:] = \textrm{DeformLayer}(\hat{F}[s,:],\{\mathcal{F}_l\},\{p_l\})$

\st $\hat{\mathcal{G}}[j,:] = \textrm{CrossAttention}(\hat{\mathcal{G}}[j,:],\hat{F}[s,:],\hat{F}[s,:])$

\efr 

\st $H = \textrm{sigmoid}(\mathcal{\hat{G}}\hat{F}^T)$

\st $\mathcal{H}\textrm{.append}(\phi_{sampling}(H,2^{-3})))$

\efr

\rt $\mathcal{H}$
\end{algorithmic}
\end{algorithm}

Alg. \ref{alg:multi-round-aggregation} shows the entire process of our multi-round feature aggregation: initially establish the relationship between text and pixels, and then refine these relationships through continuous iteration.

First, we select $k$ pixels with the highest similarity for each row on the multi-scale similarity matrix $S = \textrm{topk}(H,k)$, where $S$ is a matrix of dimension $n \times k$.

In the refinement phase (lines 12-24), we first update the embedded representations of the $k$ nearest image pixels each time with the text's representation based on the previous iteration (line 20-21 in Alg. 1). Next, we apply \textbf{DeformLayer} again to regenerate the multi-scale representation of pixels for the top-$k$ image positions (line 22 in Alg. 1). Notably, we re-use the multi-scale output $\{\mathcal{F}_l\}$ of \textbf{DeformLayer} from the initial stage as its input.



Subsequently, we update the representation of noun phrases using the weighted sum of the current top-$k$ image features (line 23 in Alg. 1). Below we will describe its implementation in detail.

Given a query $Q$, a key $K$ and a value $V$, through the attention  we can get $Q$'s updated weighted representation
\eqn{}{
\textrm{attn}(Q,K,V) = \textrm{softmax}\left( \frac{Q W_{q}(K W_{k})^{T}}{\sqrt{c}} \right) V W_{v}.
}
Here $W_{q}$, $W_k$ and $W_v$ represent the projection matrices, which project a row vector to the $c$-dimensional space.

We treat each row of $\mathcal{\hat{G}}$ as a query, $\hat{F}$ as key and value, and split them into $M$ blocks along the dimension of representation. The multi-head attention representation of $\mathcal{\hat{G}}[j,:]$ can be computed ($s = S[j,:]$)
$$
G[j,:] = \textrm{catcol}(\{
\textrm{attn}(\mathcal{\hat{G}}[j,\textrm{ids}(i)],\hat{F}[s,\textrm{idx}(i)],\hat{F}[s,\textrm{idx}(i)])
\}).
$$
where $j = 1,2,\cdots,n$ and $\textrm{catcol}$ represents the concatenation along the column and $\textrm{idx}(i)$ represents the column index set of the $i$-th sub-block. Then we sequentially perform addition, dropout, norm, and FFN operations on $G$ and $\mathcal{\hat{G}}$
\eqna{
\bar{G} & = & \textrm{dropout}(G + \mathcal{\hat{G}}), \\
\mathcal{\hat{G}} & = & \textrm{FFN}(\textrm{norm}(\mathcal{\hat{G}} + \bar{G})).
}

\subsection{Loss Function}
\label{model loss}

Once we have a series of predicted values $\mathcal{H}$ of the correlation coefficient of text and pixel, we can define the optimized loss function based on Binary Cross Entropy (BCE) and Dice loss according to the true value $Y$
\eqn{}{
\mathcal{L}(\mathcal{H},Y) = \sum_{i = 1}^I \lambda_{bce} \mathcal{L}_{bce}(\mathcal{H}_i,Y) + \lambda_{dice} \mathcal{L}_{dice}(\mathcal{H}_i,Y),
}
where $\lambda_{bce}$ and $\lambda_{dice}$ are the weight coefficients of the loss which are both set to $1$ in our experiments. Specifically, BCE loss is the average loss of all text-pixel pairs
\eqn{}{
\mathcal{L}_{bce}(\mathcal{H}_i,Y) = \frac{1}{nhw} \sum_{j = 1}^n \sum_{k = 1}^{hw} \textrm{CE}(Y(j,k),\mathcal{H}_i(j,k)),
}
where $\textrm{CE}$ is the cross entropy loss.

In general, the goal of BCE loss is to compute a binary classification loss for all pixels, but this loss does not consider the problem of class imbalance. To alleviate this problem, we introduce the Dice loss as an additional loss
$$
\mathcal{L}_{dice}(\mathcal{H}_i,Y) = \frac{1}{n} \sum_{j = 1}^n \left(1 - \frac{2 \sum_{k = 1}^{hw}\mathcal{H}_i(j,k)Y(j,k) }{ \sum_{k = 1}^{hw}\mathcal{H}_i(j,k) + Y(j,k) } \right).
$$


To provide sufficient intermediate supervision during the encoding stage, we follow the setup of~\cite{ding2022ppmn}, which applies the loss $\mathcal{L}$ to the predicted value $H$ of all refinement stages. For the inference, we obtain grounded results from the previous round of response maps with a threshold of 0.5.

\subsection{Discussion on Multi-round Feature Aggregation}
\label{discusstion of method}
In our task, we are given an embedded representation $t_j$ of $n$ words, and the goal is to assign all $m$ pixels $x_i$ in the image to these $n$ noun phrases. For convenience, we assume that $x_i$ and $t_j$ are located in a common vector space. We then iteratively optimize the representations of $x_i$s and $t_j$s through the objective function $\frac{1}{2} \sum_{i = 1}^m \sum_{j = 1}^n u_{ij}^2 \|x_i - t_j\|^2$, where $u_{ij}$ represents the probabilty that $x_i$ belongs to $t_j$.

\subsubsection{Solving $x_i$'s for known $t_j$'s}

We can define the following loss function
\eqna{
    \min_{u,x} && \mathcal{L}(u,x) = \frac{1}{2} \sum_{i = 1}^m \sum_{j = 1}^n u_{ij}^2 \|x_i - t_j\|^2, \\
    s.t. && \sum_{i = 1}^m u_{ij} = k,\quad u_{ij} \in \{0,1\},\quad k < n.
}
We add the constraint $\sum_{i = 1}^m u_{ij} = k$ and $k < n$ to avoid  trivial solutions. Obviously, if $m = n$, then $x_j = t_j$.

Assume that for each target point $t_j$, its $k$ closest points are $x_{r(j,1)},x_{r(j,2)},\cdots,x_{r(j,k)}$. If we fix $x_i$ to find the optimal point of $u_{ij}$,  we have
\eqn{}{
u_{ij} = 
\begin{cases}
   1,& i \in \{r(j,1),r(j,2),\cdots,r(j,k)\}, \\
   0, & \textrm{otherwise}.
\end{cases}
}

Next, if we fix $u_{ij}$, we then calculate the gradient of $\mathcal{L}(u,x)$ with respect to $x$ to get
\eqn{}{
\frac{\partial{\mathcal{L}}}{\partial{x_i}} = \sum_{j = 1}^n u_{ij}^2 x_i - \sum_{j = 1}^n u_{ij}^2 t_j.
}
Hence, we get the update formula of $x_i$
\eqn{}{
x_i = x_i - \alpha \frac{\partial{\mathcal{L}}} {\partial{x_i}} = (1 - \alpha \sum_{j = 1}^n u_{ij}^2)x_i + \sum_{j = 1}^n \alpha u_{ij}^2 t_j,
}
where $\alpha$ $(0<\alpha<1)$ is a step size. Note that the above equation is a batch processing method for $n$ targets, and its online update method for target $t_j$ can be given as
\eqn{}{
x_i = \left(1 - \alpha u_{ij}^2 \right) x_i + \alpha u_{ij}^2 t_j,
}
which can be further simplified to
\eqn{}{
x_i = \begin{cases}
    (1 - \alpha)x_i + \alpha t_j,\quad x_i \in \textrm{topk}(t_j),\\
    x_i, \quad \textrm{otherwise}.
\end{cases}
}
Here $x_i \in \textrm{topk}(t_j)$ means $i \in {r(j,1),r(j,2),\cdots,r(j,k)}$. Note that $u_{ij}$ is fixed, the optimization problem is a strictly convex optimization problem about $x$.  Therefore, setting an appropriate step size can ensure that the function value decreases after each gradient descent.

In line 21 of Alg. 1, we update $x_i$ with the following formula 
\eqn{}{
x_i = 
\begin{cases}
    f(x_i) + t_j, \quad x_i \in \textrm{topk}(t_j), \\
    x_i,\quad \textrm{otherwise}.
\end{cases}
}
Here $f(x_i)$ represents the encoded representation of $x_i$. At the same time, when we solve the nearest $k$ points from $t_j$, we exploit the predicted correlation coefficient instead of the Euclidean distance.

\subsubsection{Solving $t_i$'s for known $x_i$'s}
Drawing the idea of fuzzy K-means, we define the following loss function
\eqna{
    \min_{u,t} && \mathcal{L}(u,t) = \frac{1}{2} \sum_{i = 1}^m \sum_{j = 1}^n u_{ij}^2 \|x_i - t_j\|^2, \\
    s.t. &&  \sum_{j = 1}^n u_{ij} = 1,\quad i = 1,2,\cdots,m,
    }
whose Lagrangian function is
$$
\mathcal{J}(u,t,\lambda) = \frac{1}{2} \sum_{i = 1}^m \sum_{j = 1}^n u_{ij}^2 \|x_i - t_j\|^2 + \sum_{i = 1}^m \lambda_i \left( \sum_{j = 1}^n u_{ij} - 1 \right).
$$
According to KKT, we obtain the optimal $u_{ij}$ and $t_j$ satisfying 
\eqna{
u_{ij} & = & \frac{1/\|x_i - t_j\|^2}{\sum_{k = 1}^n 1/\|x_i - t_k\|^2}, \\
t_j & = & \frac{\sum_{i = 1}^n u_{ij}^2 x_i}{\sum_{i = 1}^n u_{ij}^2}.
}
The above updating formulas show that $t_j$ is the weighted average of $x_i$, and the weight of each item is inversely proportional to the distance, or proportional to the similarity.

In line 23 of our algorithm, we apply a multi-head attention mechanism to represent each $t_j$ as an adaptive weighting sum of top-k $x_i$, where the weight of $x_i$ with respect to $t_j$ is expressed as a normalized dot product.

\section{Experiments}
\subsection{Dataset and Evaluation Criteria}

We compare the performance of our proposed method with the other methods on the benchmark PNG dataset, which matches noun phrase annotations in the Localized Narrative dataset \cite{pont2020connecting} with panoptic segmentation annotations in the MS COCO dataset \cite{lin2014microsoft}. As the only publicly available benchmark in PNG, this dataset contains 726,445 noun phrases matched to segments involving 659,298 unique segments, and it covers 47.5\% of the segmentation annotations in the MS COCO panoptic segmentation dataset and 45.1\% of the noun phrases in the Localized Narrative dataset. On average, each title in the dataset has 5.1 noun phrases. The train and validation splits contains 133,103 and 8533 localized narratives, respectively.

We adopt the average recall as the evaluation metric for model performance following the previous practice. It calculates the recall for different intersection over (IoU) thresholds between the segmentation result and the ground truth, then draws curves based on different thresholds. The area under the curve represents the average recall value, i.e., for plural noun phrases. All ground truth annotations are merged into a single segment to compute the IoU result.

\subsection{Implementation Details}

Our backbone configuration is consistent with the PPMN baseline model \cite{ding2022ppmn}, where we utilize official pre-trained \cite{wu2019detectron2} ResNet101 model (with 3x schedule) on the MS COCO dataset \cite{lin2014microsoft} as the image backbone. For the text input, we use the pre-trained ``base-uncased" BERT model \cite{devlin2018bert} to convert each word in the narrative captions into a 768-dimensional vector. The longest caption contains 230 characters, with up to 30 different noun phrases that need be localized. We do not update the image and text pre-trained backbone models during training.

Furthermore, we only apply image size augmentation to the input image, which is resized to a resolution between 800 and 1,333 pixels while maintaining the aspect ratio. We implement our proposed model using PyTorch and train it with a batch size 10 for 20 epochs on three NVIDIA 3090 GPUs. The Adam optimizer is used with a fixed learning rate of $10^{-4}$. During inference, we obtain segmentation results following the configuration of the two-stage model \cite{gonzalez2021panoptic}, which averages the matching graphs of all words in each noun phrase.

\begin{figure*}[htbp]
	
	\begin{minipage}{0.30\linewidth}
		\vspace{1pt}
		\centerline{\includegraphics[width=\textwidth]{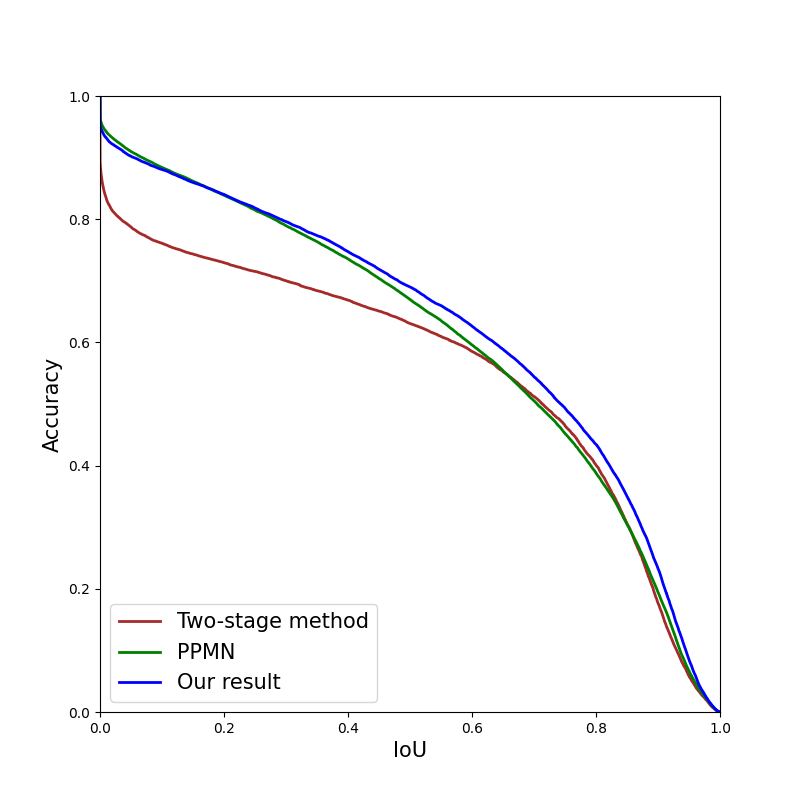}}
		\subcaption{Overall performance}
            \label{curves_a}
	\end{minipage}
	\begin{minipage}{0.30\linewidth}
		\vspace{1pt}
		\centerline{\includegraphics[width=\textwidth]{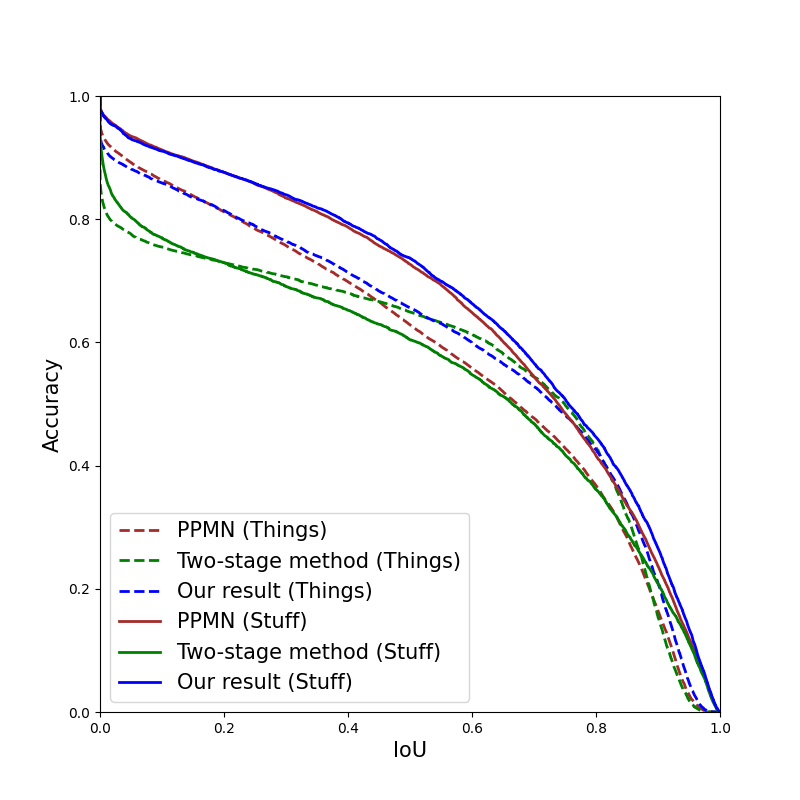}}
		\subcaption{Things and stuff performance}
            \label{curves_b}
	\end{minipage}
	\begin{minipage}{0.30\linewidth}
		\vspace{1pt}
            \centerline{\includegraphics[width=\textwidth]{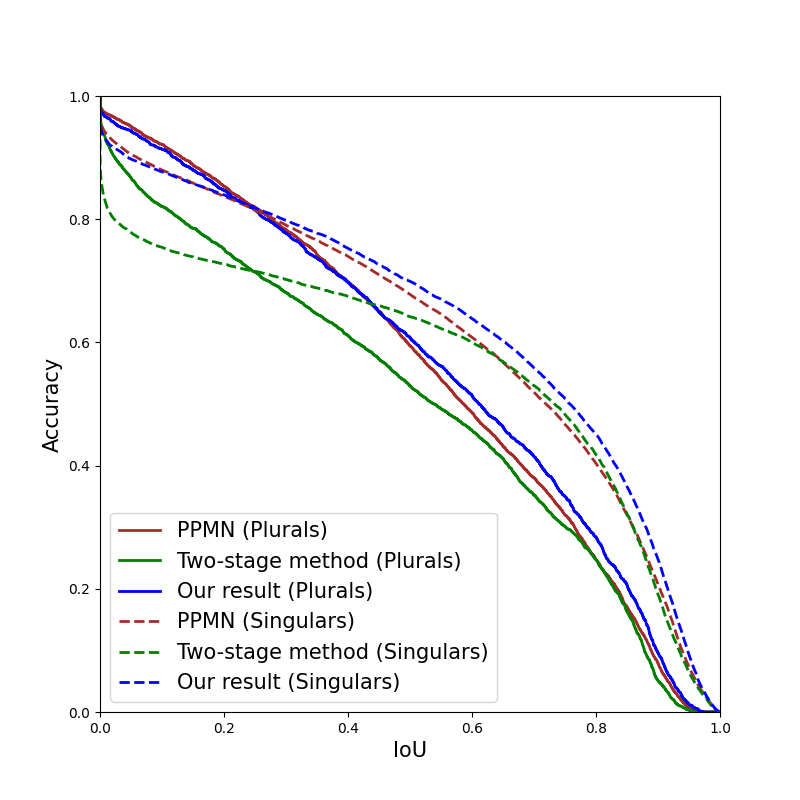}}
		\subcaption{Singulars and plurals performance}
            \label{curves_c}
	\end{minipage}
 
	\caption{Average recall curves for the PNG dataset: (a) overall performance compared to other state-of-the-art methods, (b) curves for things and categories of things, and (c) curves for singular and plural noun phrases.}
	\label{curves}
\end{figure*}

\begin{table*}[htbp]
\caption{Results of our method for the panoptic narrative base task, compared with state-of-the-art methods.}
\begin{center}
\begin{tabular}{|l|c|c|c|c|c|}
\hline
\textbf{Method}&\multicolumn{5}{|c|}{\textbf{Average Recall}} \\
\cline{2-6} 
\textbf{model} & \textbf{\textit{overall}}& \textbf{\textit{singulars}}& \textbf{\textit{plurals}}& \textbf{\textit{things}}& \textbf{\textit{stuff}} \\  
\hline
PNG \cite{gonzalez2021panoptic} & 55.4 & 56.2 & 48.8 & 56.2 & 54.3 \\
\hline
PPMN\dag \cite{ding2022ppmn} & 56.7 & 57.4 & 49.8 & 53.4 & 61.1 \\
\hline
EPNG \cite{wang2023towards} & 58.0 & 58.6 & 52.1 & 54.8 & 62.4 \\
\hline
PPMN \cite{ding2022ppmn} & 59.4 & 60.0 & 54.0 & 57.2 & 62.5 \\
\hline
DRMN(Our) & \textbf{62.9} \color{red}(+3.5) & \textbf{63.6} \color{red}(+3.6) & \textbf{56.7} \color{red}(+2.7) & \textbf{60.3} \color{red}(+3.1) & \textbf{66.4} \color{red}(+3.9) \\
\hline
\end{tabular}
\label{tab:main-results}
\end{center}
\end{table*}

\subsection{Experimental Results}


To validate the effectiveness of the context information we introduced, we compare the performance of our proposed model with other methods on the PNG dataset. The main results are shown in Table \ref{tab:main-results}. We also compare the recall curves of these methods in Fig.~\ref{curves}. It is worth noting that our best model do not update the image and text backbones during training, and the results obtained using the same training strategy are labeled with PPMN\dag.


Compared to the current state-of-the-art methods on the PNG dataset, our proposed model achieves an average $3.5\%$ (from 2.7\% to 3.9\%) improvements in average recall across various metrics. Specifically, our method achieves 3.5/3.1/3.9/3.6/2.7 improvements in whole/thing/thing/singular/plural categories, validating the effectiveness of our proposed method.


Fig.~\ref{curves} depicts recall values for different classes at different IOU thresholds. In Fig.~\ref{curves_a}, when the IoU threshold is larger than 0.3, our method (blue curve) consistently outperforms the baseline model (green curve), showing that the image context information can indeed  benefit the segmentation results. Furthermore, in Fig.~\ref{curves_b}, we can see that our method exhibits significant performance gains in object categories compared to the baseline model, even approaching the accuracy of the two-stage method. This further demonstrates that  context information could enhance the representation ability of the aggregated text feature, leading to better segmentation results. In Fig.~\ref{curves_c}, we further investigate the detailed performance of our method on object categories (stuff and singular):  our method still improves the segmentation results on both categories, which indicates that the essential context information  may benefit all object categories' results.


\subsection{Ablation Studies}


To validate the effectiveness of our proposed method on different components, we conduct ablation experiments on the PNG task and compare the results under different parameter settings.

\subsubsection{Number of deformable encoder layer}

In Table~\ref{tab:encoding-layer}, we show how  various deformable encoder layers affect the model's performance. The results show that combining multi-scale image context information can improve segmentation performance, whereas  too many encoder layers may lead to performance degradation. 


\begin{table}[htbp]
\caption{Ablation study on the number of encoder layers.}
\begin{center}
\begin{tabular}{|c|c|c|c|c|c|}
\hline
\textbf{Num of Encoder}&\multicolumn{5}{|c|}{\textbf{Average Recall}} \\
\cline{2-6} 
\textbf{Layers} & \textbf{\textit{overall}}& \textbf{\textit{singulars}}& \textbf{\textit{plurals}} & \textbf{\textit{things}}& \textbf{\textit{stuff}} \\
\hline
0  & 61.5 & 62.2 & 55.4 & 58.7 & 65.4 \\
\hline
1 & 62.4 & 62.9 & 57.1 & 60.1 & 65.6 \\
\hline
2 & \textbf{62.9} & \textbf{63.6} & \textbf{56.7} & \textbf{60.3} & \textbf{66.4} \\
\hline
3 & 62.6 & 63.3 & 55.8 & 60.2 & 65.9 \\
\hline
\end{tabular}
\label{tab:encoding-layer}
\end{center}
\end{table}

\subsubsection{Number of rounds for multi-round feature aggregation}


We also examine the performance of different rounds of feature aggregation on the model in Table \ref{tab:refined-layers} where the results show that introducing multi-round feature aggregation suddenly improves model performance. We also observe that the performance of the Singular and Stuff categories gradually improves as the number of stages increases. However, there are some fluctuations in the performance of the plurals category. Since we find  some incomplete annotations in the PNG dataset for the plurals category during training (as shown in Fig. \ref{vis_result_2}), we believe it is reasonable for the model to be slightly unstable in testing on this category.

\begin{table}[htbp]
\caption{Ablation study on the number of feature aggregation round.}
\begin{center}
\begin{tabular}{|c|c|c|c|c|c|}
\hline
\textbf{Number of}&\multicolumn{5}{|c|}{\textbf{Average Recall}} \\
\cline{2-6} 
\textbf{Rounds} & \textbf{\textit{overall}}& \textbf{\textit{singulars}}& \textbf{\textit{plurals}} & \textbf{\textit{things}}& \textbf{\textit{stuff}} \\
\hline
0  & 59.6 & 60.2 & 54.5 & 57.1 & 63.0 \\
\hline
1  & 62.7 & 62.7 & \textbf{57.2} & \textbf{60.4} & 65.9 \\
\hline
2 & 62.7 & 63.5 & 56.2 & 60.1 & 66.4 \\
\hline
3 & \textbf{62.9} & \textbf{63.6} & 56.7 & 60.3 & \textbf{66.4} \\
\hline
\end{tabular}
\label{tab:refined-layers}
\end{center}
\end{table}

\subsubsection{Number of sample points for multi-round feature aggregation module}


We conduct  further studies to evaluate  our proposed multi-round feature aggregation module by examining the impact of different numbers of sampling points on model performance. The results are reported in Table \ref{tab:sample-points}. Since the context information convers a large extent on image during top-k image refinement stege, as further shown in Fig.~\ref{fig:sampling-layer}, our model can perform well even with a small number of sample points, This indicates a small set for the context information refined top-$k$ image features may be enough to cover the object information. We observe that increasing the number of sampling points improves the stuff category more. We attribute this to the mask of stuff category which usually covers more space in the image. Hence increasing the sampling points may help to preserve the semantics of different parts of the ground truth mask information.


\begin{table}[htbp]
\caption{Ablation on the number of sampled image points.}
\begin{center}
\begin{tabular}{|c|c|c|c|c|c|}
\hline
\textbf{Sampling}&\multicolumn{5}{|c|}{\textbf{Average Recall}} \\
\cline{2-6} 
\textbf{Points ($k$)} & \textbf{\textit{overall}}& \textbf{\textit{singulars}}& \textbf{\textit{plurals}} & \textbf{\textit{things}}& \textbf{\textit{stuff}} \\
\hline
10 & 62.8 & 63.4 & 56.9 & 60.4 & 66.1 \\
\hline
50 & \textbf{62.9} & \textbf{63.6} & \textbf{56.7} & \textbf{60.4} & \textbf{66.4} \\
\hline
100 & 62.9 & 63.6 & 56.7 & 60.3 & 66.4 \\
\hline
400 & 62.7 & 63.3 & 56.5 & 60.2 & 66.2 \\
\hline
\end{tabular}
\label{tab:sample-points}
\end{center}
\end{table}

\subsection{Qualitative Analysis}
\label{Qualitative Analysis}



\begin{figure}[htbp]
\centerline{\includegraphics[width=.95\linewidth]{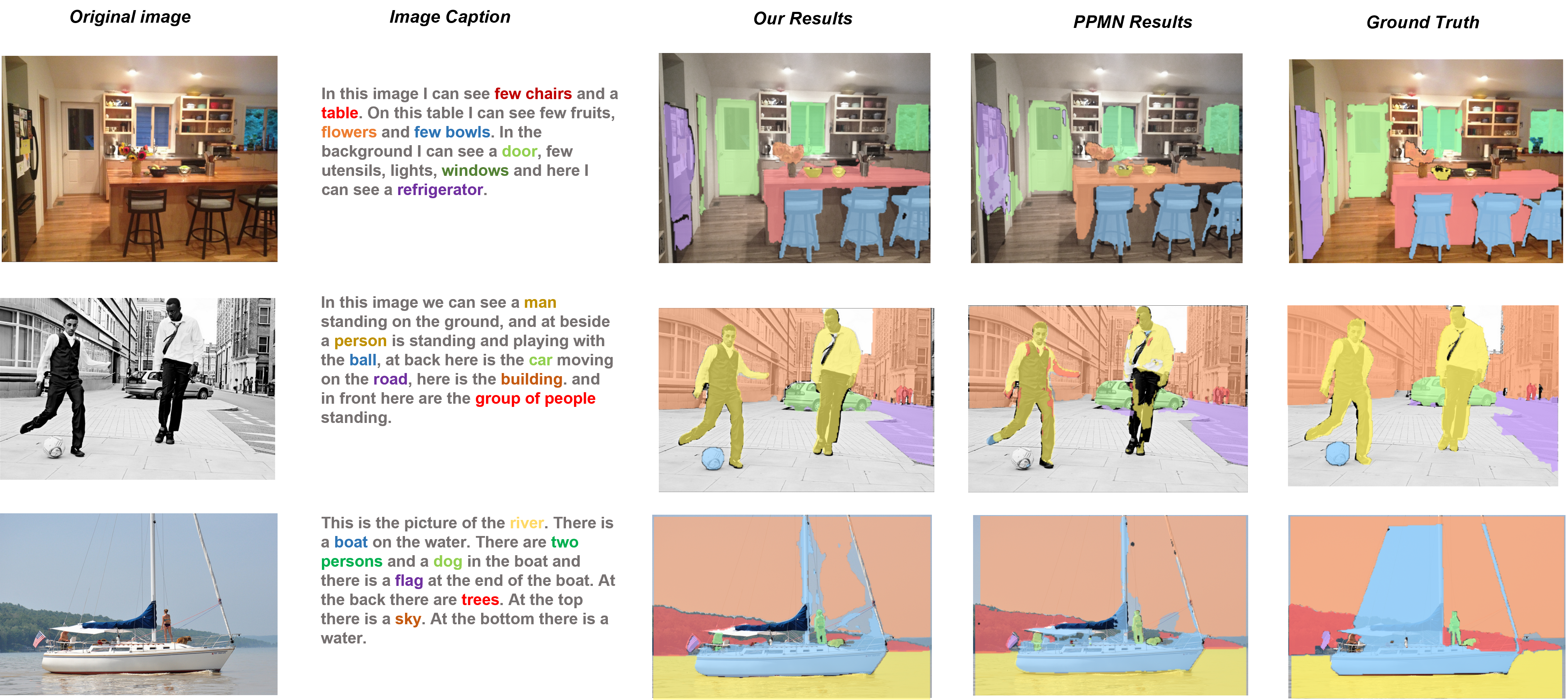}}
\caption{Qualitative results for Panoptic Narrative Grounding. The segmentation masks in the image correspond one-to-one to the colors mentioned in the text.}
\label{vis_result_2}
\end{figure}

We illustrate the qualitative results of our proposed model for text paragraphs in Fig. \ref{vis_result_2}. It is observed that comparing to the baseline model, our model predicts more complete segmentation results (``doors", ``windows" and ``refrigerator" in the fist row, ``person", ``group of people" in the second row), indicating that  refinement of top-$k$ sampled image features benefits to cluster more related pixels. It is worth mentioning that the model is even able to locate more complete ground truth annotation (the ``few bowls" result in the first row, the ``two persons" result in the third row).


\begin{figure}[htbp]
\centerline{\includegraphics[width=.95\linewidth]{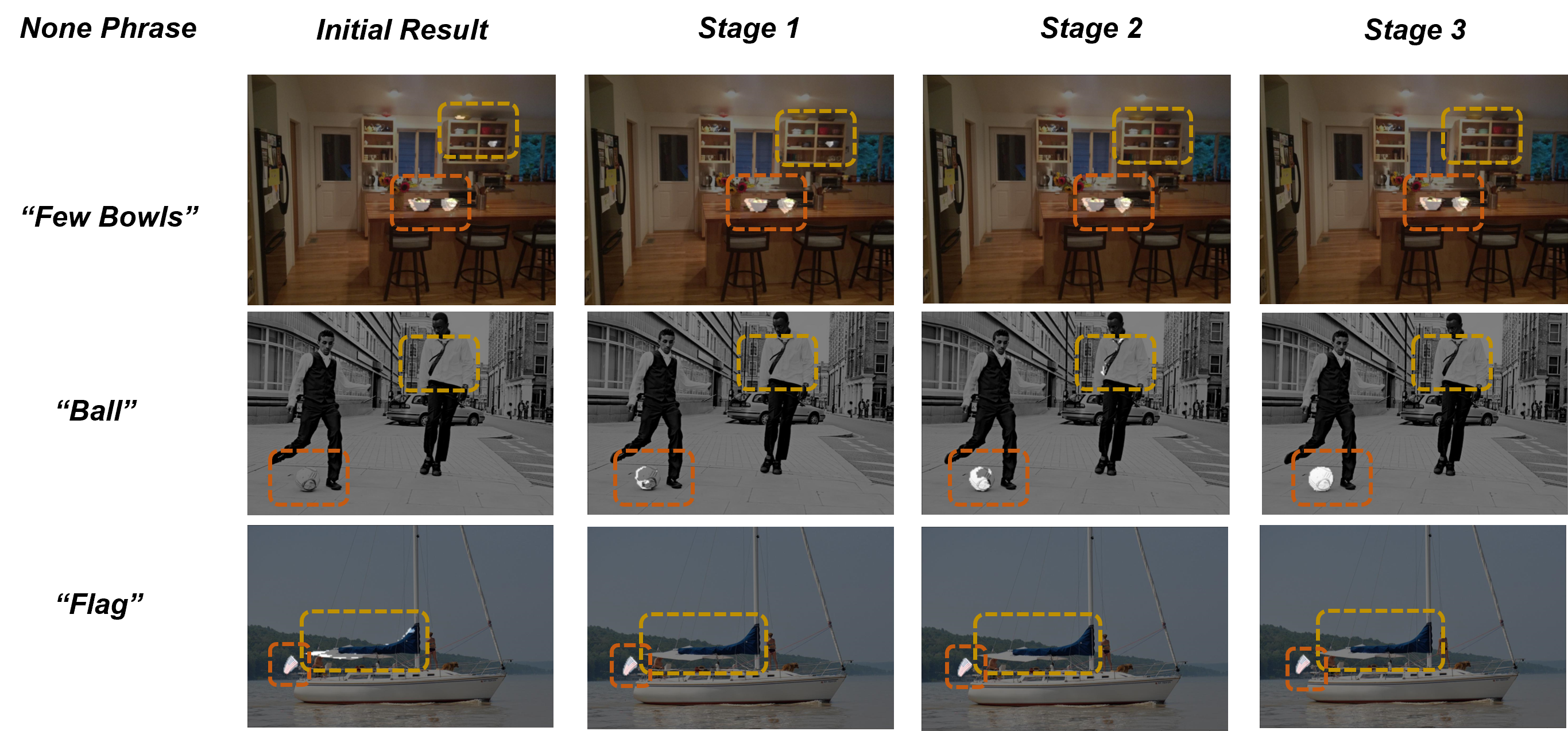}}
\caption{Refinement results in each stage for specific visual objects. Distracting objects and target objects are highlighted in dashed yellow and red boxes, respectively.}
\label{vis_result_1}
\end{figure}


To visualize how the context information benefits segmentation, we show the instructive results during the multi-round feature aggregation process in Fig.~\ref{vis_result_1}. In the first and third rows, the model gradually filters out irrelevant results in each round of refinement, improving segmentation results despite irrelevant or similar objects in the initial matching. The example in the second row shows how a segmented object goes from a low response matching result to an almost correct matching during refinement. Compared to the example in Fig.~\ref{problem_example}, such results validate that context information does matter in alleviate the phrase-to-pixel mis-match and thus improve the performance in PNG.

\begin{figure}[htbp]
\centerline{\includegraphics[width=.95\linewidth]{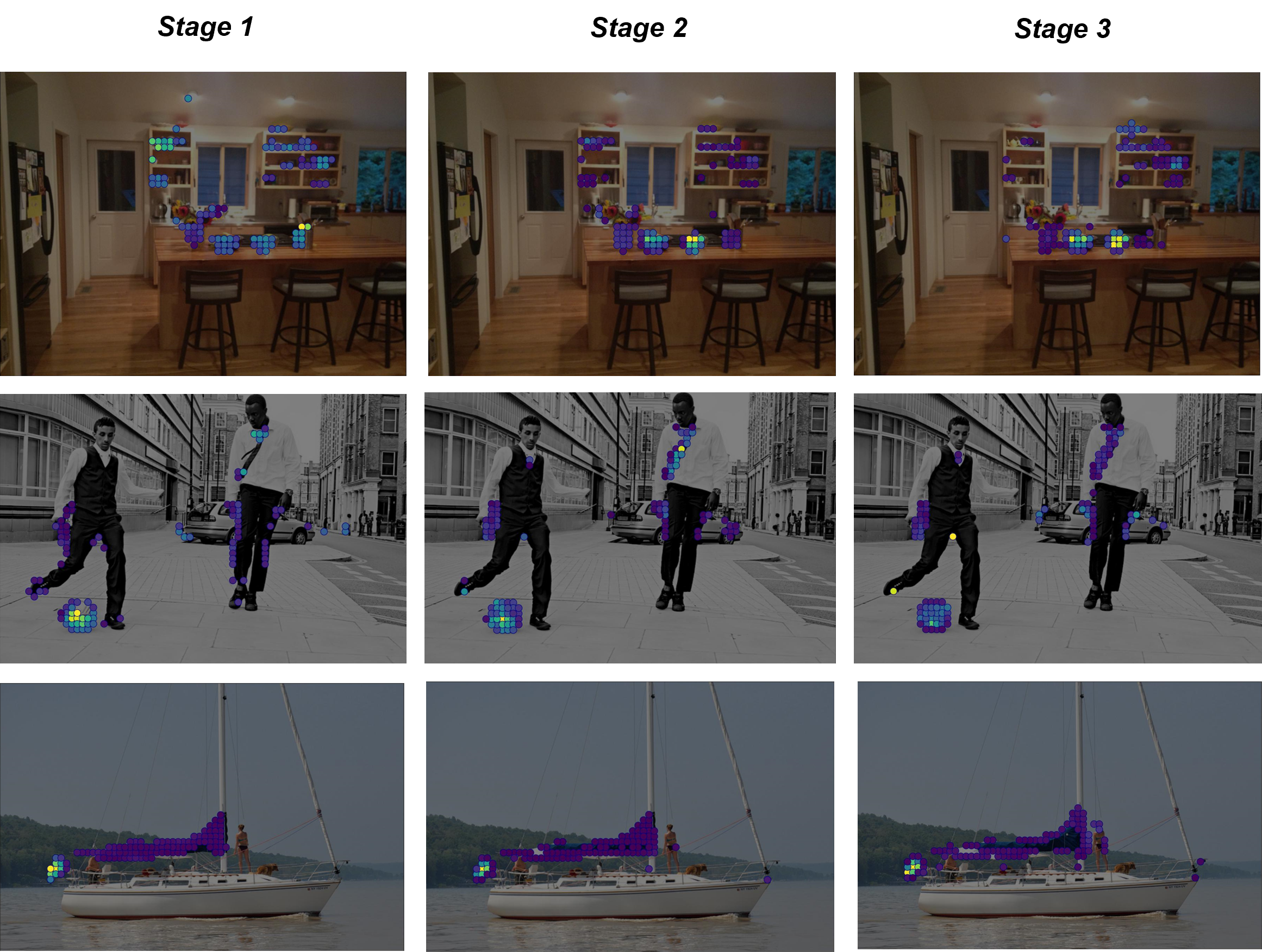}}
\caption{Attention weights for top-$k$ image locations. Weights are averaged over all heads in a multi-head cross-attention layer. Lighter colors indicate greater weight and vice versa.}
\label{fig:topk}
\end{figure}

We further visualize the top-$k$ image locations most similar to the text during each round of refinement and corresponding weights in the cross-attention mechanism in Fig.~\ref{fig:topk}. As seen in the figure, our proposed method generally puts the weight on the most relevant objects and gradually filters out the impact of irrelevant objects.

\begin{figure}[htbp]
\centerline{\includegraphics[width=.95\linewidth]{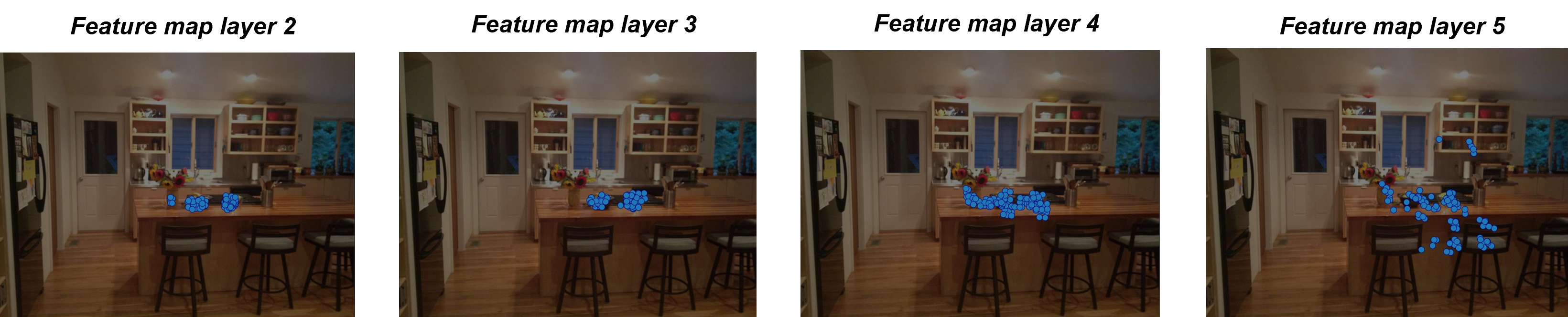}}
\caption{Visualization of the most important offset points in the deformable attention layer, according to the top-$k$ query of phrase ``few bowel". We visualize the top-50 most important points based on the attention weights for each layer of the multi-scale feature map.}
\label{fig:sampling-layer}
\end{figure}

To better visualize what context information is introduced by deformable attention during the aggregation stage, we further visualize the 50 most important offset points obtained by sampling the top-$k$ most relevant points in the last round of the deformable attention mechanism for different layers in Fig. \ref{fig:sampling-layer}. { We can find that in the primary layers containing relatively detailed low-level information, the offset point is around the target object, which may refine more detailed information for the target object. In the latter layers that contain relatively high-level information, the offset points attend to the general context around the target object. The above observations indicate our proposed refinement method effectively concentrates on both detailed and general context information for the top-$k$ sampled image points.}


\section{Conclusion}



In this paper, we propose a novel one-stage model named Deformable-Attention Refined Matching Network (DRMN) for Panoptic Narrative Grounding (PNG) task. Built upon the end-to-end one-stage model architecture, we integrate the essential context information of multi-scale image features in the multi-modal information fusion module as an addition cue to enhance the feature discriminative ability. Furthermore, we employ a clustering framework to interpret our proposed module and validate our method through experiments on the benchmark PNG dataset. The results demonstrate that our proposed model can achieve new  state-of-the-art performance, with a 3.5\%  improvement on the average recall metric.

\bibliographystyle{IEEEtran}
\bibliography{reference/ref}{}

\end{document}